\documentclass[twoside,leqno,twocolumn]{article}

\usepackage[letterpaper]{geometry}

\usepackage{ltexpprt}
\usepackage{hyperref}

\usepackage{cite}

\usepackage{amsmath,amssymb,amsfonts,amsthm}

\usepackage{algorithmic}
\usepackage[algo2e, ruled,vlined]{algorithm2e}
\usepackage{graphicx}
\usepackage{textcomp}
\usepackage{xcolor}
\usepackage{mathtools}
\usepackage{subfig}
\usepackage{lipsum}

\theoremstyle{definition}
\newtheorem{definition}{Definition}[section]

\makeatletter
\def\blfootnote{\xdef\@thefnmark{}\@footnotetext}
\makeatother

\begin{document}
\newcommand\relatedversion{}
\renewcommand\relatedversion{\thanks{The full version of the paper can be accessed at \protect\url{https://arxiv.org/abs/1902.09310}}} 

\title{\Large HintNet: Hierarchical Knowledge Transfer Networks for Traffic Accident Forecasting on Heterogeneous Spatio-Temporal Data \vspace{-0.1in}}

\author{Bang An\thanks{University of Iowa, \{bang\-an, xun\-zhou, nick\-street\}@uiowa.edu}
\and Amin Vahedian\thanks{Northern Illinois University, avahediankhezerlou@niu.edu}
\and Xun Zhou$^*$
\and W. Nick Street$^*$
\and Yanhua Li\thanks{Worcester Polytechnic Institute, yli15@wpi.edu}
}

\date{}

\maketitle


\fancyfoot[R]{\scriptsize{Copyright \textcopyright\ 2022 by SIAM\\
Unauthorized reproduction of this article is prohibited}}




\vspace{-0.3in}
\begin{abstract} \small\baselineskip=9pt Traffic accident forecasting is a significant problem for transportation management and public safety. However, this problem is challenging due to the spatial heterogeneity of the environment and the sparsity of accidents in space and time. The occurrence of traffic accidents is affected by complex dependencies among spatial and temporal features. Recent traffic accident prediction methods have attempted to use deep learning models to improve accuracy. However, most of 
these methods either focus on small-scale and homogeneous areas such as populous cities or simply use sliding-window-based ensemble methods, which are inadequate to handle heterogeneity in large regions. To address these limitations, this paper proposes a novel Hierarchical Knowledge Transfer Network (\textbf{HintNet}) model to better capture irregular heterogeneity patterns. HintNet performs a multi-level spatial partitioning to separate sub-regions with different risks and learns a deep network model for each level using spatio-temporal and graph convolutions. Through knowledge transfer across levels, HintNet archives both higher accuracy and higher training efficiency. Extensive experiments on a real-world accident dataset from the state of Iowa demonstrate that HintNet outperforms the state-of-the-art methods on spatially heterogeneous and large-scale areas.\end{abstract}
\blfootnote{Xun Zhou is the corresponding author}
\vspace{-0.1in}
\section{Introduction}
\label{intro}
    
Traffic accident is a major safety concern in modern society. In the United States, it is estimated that nearly 40 thousand people died in traffic accidents in 2020, according to the National Highway Traffic Safety Administration (NHTSA) \cite{2020fatality}. This number shows an increase of around 7\% compared to 2019, despite the effects of the coronavirus pandemic on mobility. Moreover, the US Census Bureau found that the average commuting time of Americans has increased by 10\% between 2006 and 2019 \cite{burd2021travel}. With nearly 85\% of Americans driving for their commute \cite{burrows2021commuting} and the increased average commute time, the number of fatalities can continue to grow in the coming years. Therefore, the ability to forecast accidents is of significant importance to the members of society, as it can help trigger public safety preparedness as well as alert drivers to the potential risk of accidents at certain locations at certain times in the future. The availability of large-scale data on accidents, climate, road dynamics, and other spatial and temporal characteristics has enabled the researchers to propose machine learning approaches to traffic accidents.

Accidents are a relatively rare phenomenon. Of all the location-time instances, very few of them contain a traffic accident. As a result, it is challenging for a machine learning method to learn the complex patterns leading up to an accident, with the presence of an overwhelming majority of times and locations that have zero accidents. Moreover, such patterns are not likely to be homogeneous over space and time. The patterns that predict traffic accidents in a crowded urban area likely differ from such patterns in rural areas. In other words, the patterns that may predict traffic accidents are heterogeneous. Given how challenging it is to learn such patterns in the first place (due to the overwhelming imbalance between accident, no-accident), it is even more challenging to learn all these spatially heterogeneous patterns in the same model. 

Researchers have tackled the accident prediction problem with a variety of approaches. A large body of literature has explored solutions with non-machine learning methods, or has proposed straightforward applications of data mining techniques to predict accidents \cite{chang2005analysis,chang2005data,abellan2013analysis,caliendo2007crash,bergel2013explaining,eisenberg2004mixed}. Such techniques do not address the challenges of sparsity and heterogeneity. More recently, researchers have proposed deep learning techniques \cite{chen2016learning, HuangChao2019DDFN} and have taken advantage of LSTM methods and attention mechanisms to learn temporal patterns, and have used convolutional methods to learn local spatial patterns \cite{najjar2017combining}. Wang et al. \cite{wang2021gsnet} proposed GSNet to capture multi-scale spatial-temporal dependencies by applying a geographical module and a semantic module. Zhou et al. \cite{zhou2020riskoracle} handled the sparsity issue using a transformation strategy to discriminate the risk values, dominated by zero or near-zero values. None of these methods directly address the spatial heterogeneity with mixed urban and rural areas, and only rely on the model to learn such patterns from spatial features. Notably, Yuan et al. \cite{yuan2018hetero} proposed a Hetero-ConvLSTM model to predict traffic accidents in a grid setting. Their method addresses the heterogeneity challenge by an ensemble of predictions from 21 pre-selected sub-regions in a sliding-window manner. However, the pre-selected windows could not fully capture heterogeneity as they ignore the underlying spatial patterns. Moreover, it does not utilize the shared knowledge across windows and has a high computational cost due to the ensembles.  


In this paper, we propose \textbf{HintNet}, a \underline{\textbf{Hi}}erarchical K\underline{\textbf{N}}owledge \underline{\textbf{T}}ransfer \underline{\textbf{Net}}work model to address the spatial heterogeneity problem.  HintNet first employs a hierarchical spatial partitioning method to systematically group regions with potentially similar risk patterns together without needing prior knowledge of the regions. 
Then, separate models for each level of the hierarchy are trained, allowing the unique patterns of each level to be learned separately. Moreover, we argue that despite heterogeneity, there is also a pattern of traffic accidents that is common among all levels. We develop a knowledge transfer method to allow the models to share knowledge across the different levels of the hierarchy. We take advantage of the expansive set of measured and derived features from numerous datasets from the state of Iowa to train HintNet. Our extensive evaluations show that HintNet is successful in outperforming the state-of-the-art baselines. Moreover, our evaluations show that our proposed knowledge transfer mechanism can improve the prediction accuracy and shorten the training process of the models significantly. Our contributions are as follows:
\begin{itemize}
    \vspace{-0.07in}
    \item We propose a hierarchical space partitioning framework to automatically group regions with potentially different accident patterns.
    \vspace{-0.07in}
    \item We propose a deep neural network to predict the traffic accidents of a region based on spatial, temporal and spatio-temporal features with jointly-trained graph convolution and LSTM modules.
    \vspace{-0.07in}
    \item We propose a knowledge transfer mechanism to share the common pattern of accidents across regions among all the models to expedite the training and improve accuracy.
    \vspace{-0.07in}
\end{itemize}

The rest of the paper is organized as follows: In the next section, we discuss the related work. Next, we present an overview of the dataset and extracted features as well as preliminary concepts and a problem formulation. Then, we present our solution, HintNet, followed by the experiments and the conclusion.

\vspace{-0.1in}
\section{Related Work}
\label{related}

A large body of work has taken a straightforward approach of simply applying existing prediction models to the accident prediction problem including ANNs and decision trees \cite{chang2005analysis,chang2005data}, random forest ensemble\cite{lin2015novel}, or Probabilistic Neural Networks \cite{abellan2013analysis}. Caliendo et al. developed multiple regression models based on Poisson, Negative Binomial, and Multinomial analyses to predict the number of accidents in given roads \cite{caliendo2007crash}. Many other works used regression and correlation methods to predict the number of accidents in general and special cases \cite{bergel2013explaining,eisenberg2004mixed}. However, these works only applied existing methods to the problem, and do not offer methodological contributions that address the challenges specific to the traffic accident prediction problem.

The use of deep learning\cite{chen2016learning,najjar2017combining} to predict traffic accidents is relatively recent. However, most of these early attempts only use data from a single view (e.g., spatial or temporal) for prediction, therefore unable to fully capture the spatiotemporal patterns. Wang et al. \cite{wang2021gsnet} proposed GSNet with a geographical module and a semantic module to capture a diverse set of features to learn the patterns. The geographic features aim to allow the model to learn the heterogeneity in space. However, their method does not fully address the heterogeneity issue, as they are still relying on the model to capture it through learning patterns in the features. Zhou et al. \cite{zhou2020riskoracle} offers a novel transformation of zero-accident instances to handle the sparsity issue. However, they also do not fully address the heterogeneity problem and rely on a single model to learn the spatial heterogeneity directly from features, even though the number of samples with accidents is limited due to sparsity. Yuan et al. proposed Hetero-ConvLSTM~\cite{yuan2018hetero}, which explicitly addresses spatial heterogeneity in accident prediction. This deep learning method divides the spatial field into 21 distinct regions. They then build an ensemble method to predict the number of accidents. This method addresses the temporal patterns and spatial heterogeneity. However, the manually-selected regions do not necessarily reflect different prediction patterns.

Distinct from all the above methods, our HintNet method fully addresses the spatial heterogeneity problem using an automatically generated hierarchical partitioning of the space, a deep learning network with spatial, temporal, and spatio-temporal features and using a knowledge transfer framework to train a diverse set of models to capture the heterogeneous patterns of accidents in space and time.

\vspace{-0.1in}
\section{Preliminary Concepts and Overview}
In this section, we introduce the data sources, extracted features, and our problem formulation. 
Our data covers the entire state of Iowa, which is a suitable place to study traffic accident forecasting problem due to the heterogeneous environment with both rural and urban areas. 

\begin{figure}[t]
    \centering
    
    \begin{minipage}{0.225\textwidth}
    \includegraphics[width=1\textwidth]{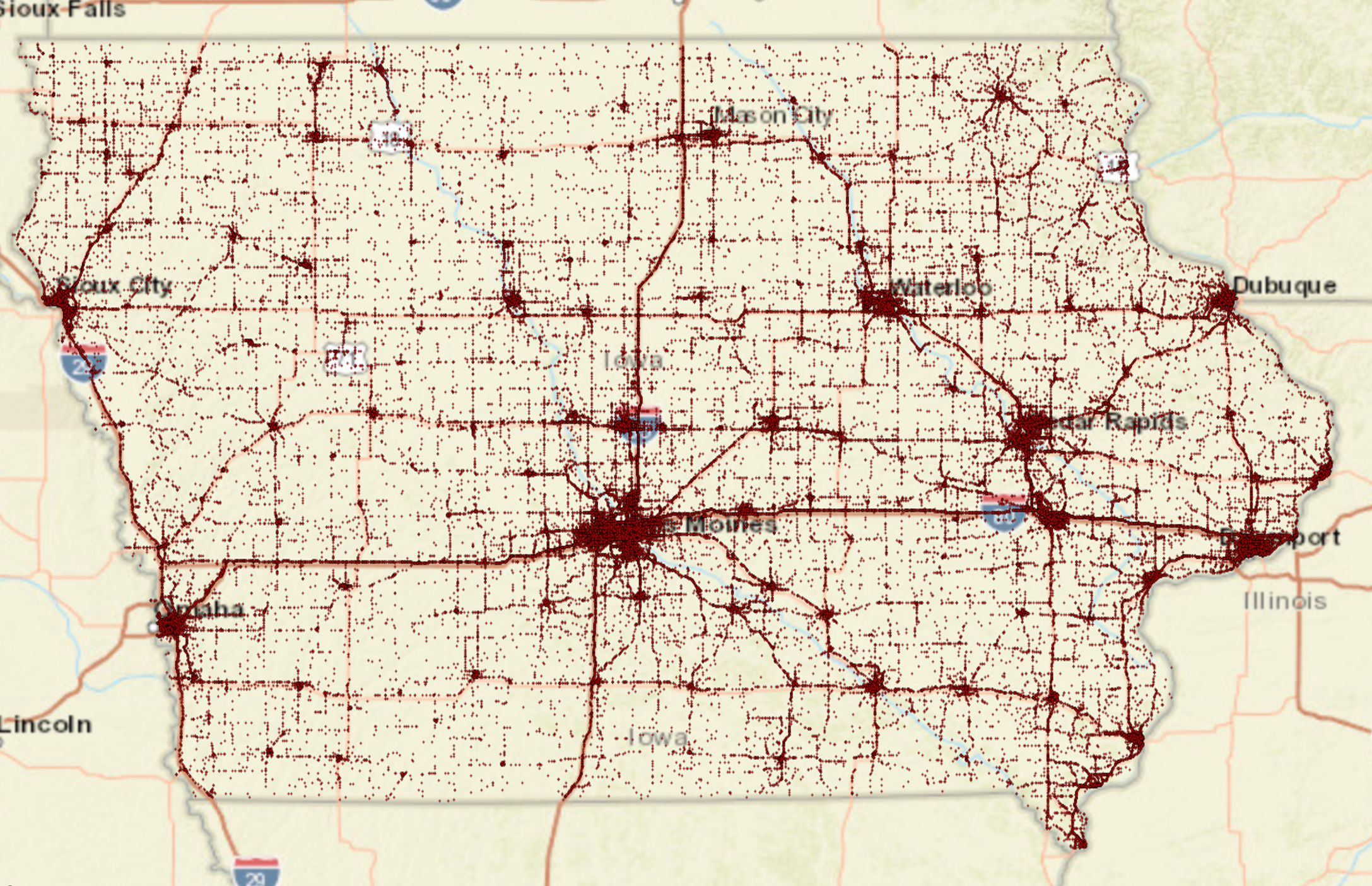}\hfill
    \caption*{\footnotesize {(a) traffic accidents in Iowa}}
    \end{minipage}\hfill
    \begin{minipage}{0.225\textwidth}
    \includegraphics[width=1\textwidth]{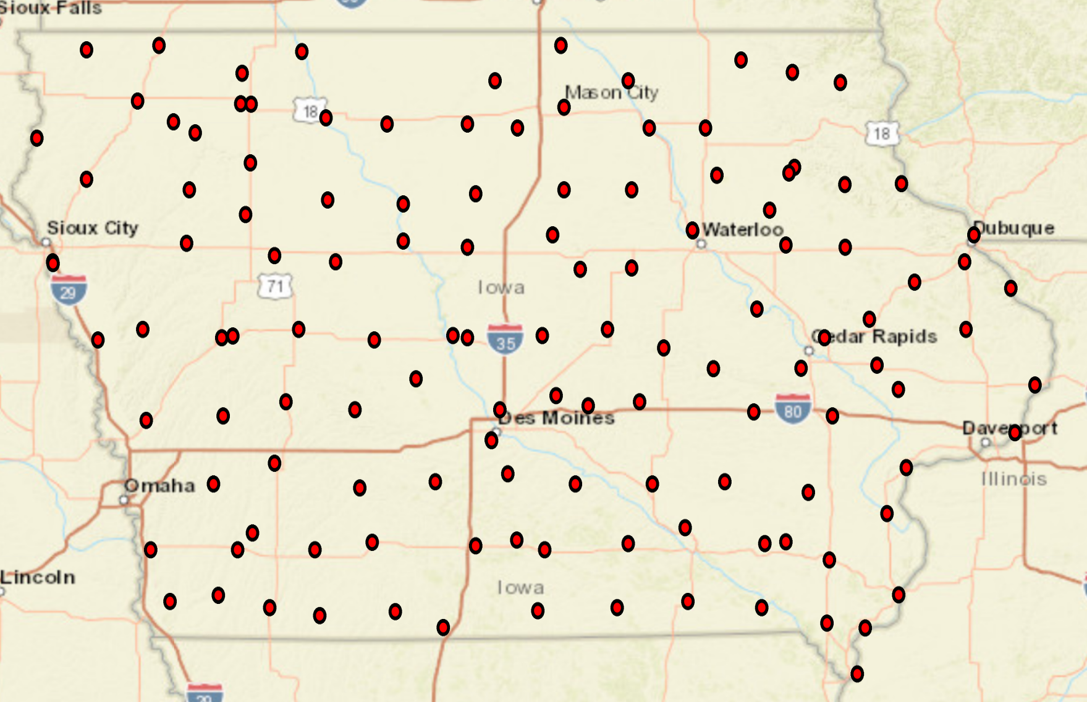}
    \caption*{\footnotesize{(b) COOP Stations}}
    \end{minipage}\hfill
\vspace{-0.1in}
	\caption{\footnotesize{Visualization of traffic accidents and COOP data}}
	\vspace{-0.2in}
	\label{three_data_map}
\end{figure}
\vspace{-0.15in}
\subsection{Data Sources}
The data we collected are all within the range of Iowa from the year 2016 to 2018. We collect data from the following sources: \textbf{(1) Vehicle Crash data} is collected by the Iowa Department of Transportation(DOT)\footnote{\tiny{ \url{https://icat.iowadot.gov/\#}}}. The data contains the 168,964 crash records from the year 2016 to 2018. The records include the time and location of each crash. Figure \ref{three_data_map} (a) shows the mapping of the traffic accident records in the state of Iowa. \textbf{(2) RWIS (Roadway Weather Information System)}\footnote{\tiny{ \url{https://mesonet.agron.iastate.edu/RWIS/}}} is an atmosphere monitoring system with 86 observation stations located at the state primary roads. \textbf{(3) COOP (National Weather Service Cooperative Observer Program)}\footnote{\tiny{ \url{https://mesonet.agron.iastate.edu/request/coop/obs-fe.phtml}}} is maintained by National Weather Service to monitor weather information. Unlike RWIS, COOP concentrates on weather data such as precipitation, snowfall, and snow depth. Figure \ref{three_data_map} (b) demonstrates the locations of the observation stations. \textbf{(4) POI.} The Point-of-Interest data are collected from HERE MAP API\footnote{\tiny{ \url{https://developer.here.com/documentation/places/dev_guide/topics/categories.html}}}. We collected the 13 categories of POI with their latitude and longitude.
\textbf{(5) Iowa Road Networks.}
From Iowa DOT OPEN DATA\footnote{\tiny{ \url{https://data.iowadot.gov/datasets/f07494c9bc6048d8a34c50af400f2264l}}}, we obtained Iowa road network data with basic road information. It consists of the speed limit, estimated Annual Average Daily Traffic volume for the primary roads and secondary roads. \textbf{(6) Traffic Camera Data.}
The real-time traffic condition data were collected from 128 camera stations along state highways.
\vspace{-0.1in}
\subsection{Definitions and Feature Extraction}
Next, we define concepts needed to formulate our problem and then explain the features extracted for the prediction task. Finally, we present our problem definition. 
\begin{definition}
A spatio-temporal field $L \times T$, where $L = \{l_{1}, l_{2},...,l_{m}\}$ is a grid, where each grid cell $l_{i}$ is a $d \times d$ square area. $T = \{t_{1}, t_{2},...,t_{n}\}$ is a study period partitioned into equal time intervals (e.g., hours, days). 
\end{definition}
We map all the features and the accidents onto the grid $L$ and over time $T$. We use $C(l, t)$ to denote the total accident count in location $l$ during time $t$. Depending on the dimensions of the features, we have spatial, temporal, and spatio-temporal (ST) features, as defined later in this section. 

\begin{definition}
\textbf{Road Network Mask Map} $H$ is a binary mask layer created by mapping the road network with primary and secondary roads onto the grids. We use a spatial mask to indicate if a grid cell contains any road segments (1) or not (0). 
\end{definition}


\begin{definition}
\textbf{Temporal Features} Temporal features $F_T$ include the day of the week, day of the year, and the month of the year, whether this is a weekend, and whether this is a holiday. $F^i_T(l, t)$ represents the $i$-th temporal feature at time interval $t$ at location $l$. Note all cell locations $l$ in the study area share the same temporal features in each time interval $t$. 
\end{definition}

\begin{definition}
\textbf{Spatial Features} Spatial features $F_S$ include static features that do not change over time,  
where $F_S^i(l, t)$ represents the $i$-th spatial features for location $l$ at time interval $t$. These features include the point of interest (POI), basic road network information, and spectral features\cite{yuan2018hetero}. Specifically, every POI feature is represented by the frequency of each POI category in a grid cell $l$. Road network features include basic road conditions such as Annual Average Daily Traffic (AADT), average traffic speed, etc. To better address the spatial heterogeneity problem, we use an idea proposed by Yuan et al. \cite{yuan2018hetero} and apply the spectral analysis on the road networks to generate 10 spectral features for each grid cell, which contain spatial connectivity relationships between different locations through the road network.

\end{definition}

\begin{definition}
\textbf{Spatio-Temporal (ST) Features} Spatio-Temporal features $F_{ST}$ are those, which vary both in space and time, where $F^i_{ST}(l, t)$ represents the $i$-th ST feature for location $l$ at time interval $t$. 
$F_{ST}$ includes daily weather and traffic conditions in each location $l$ and each time slot $t$. Weather features consist of precipitation, snowfall, snow depth, etc. The weather features are continuously distributed over the entire space. The traffic condition features include average traffic speed, normal vehicle traffic volume, and truck traffic volume for each location and time interval.

\noindent\textbf{Missing value imputation}: Many ST features are only collected at sampling sites or stations. There are also missing values due to data quality issues. To utilize the data for the entire study area, we use spatial interpolation methods to impute the missing values. Specifically, we use Ordinary Kriging \cite{CressieNoel1993SfSD} to estimate the weather-related attributes for locations without a station and a Universal Kriging~\cite{CressieNoel1993SfSD} with network distance to estimate traffic-related features. 

In total, we extracted 47 features, including 29 spatial features, 5 temporal features, and 13 ST features for each grid cell $l$ and time interval $t$.
\end{definition}
\vspace{-0.2in}
\subsection{Problem Definition}
\hfill
\\
Now we are ready to define the problem formally. \\
%
\textbullet\ Given:\\
\textendash\ A spatial-temporal field $L \times T$\\
\textendash\ A road network mask map $H$ \\
\textendash\ Traffic accident count tensor $C$ for a time window $[t-n, t-1]$ for all $l\in L$, $n<t$\\
\textendash\ A set of feature tensors $F = \{F_T, F_S, F_{ST}\}$ for the same time window for all the locations $l\in L$\\
\textbullet\ Find:\\
\textendash\ Predicted accident count in every $l\in L$ for $t$: $\hat{C}(l,t)$\\
\textbullet\ Objective:\\
\textendash\ Minimize the prediction error\\
\textbullet\ Constraints:\\
\textendash\ All traffic accidents occur along road system.\\
\textendash\ Spatial heterogeneity exist in the data.

Here $n$ is the time length of the input features for each prediction. In this paper, we choose $t$ as a single day and use seven consecutive days of data to predict for the eighth day ($n=7$).

\begin{figure}
    \centering
	\includegraphics[width=0.5\textwidth]{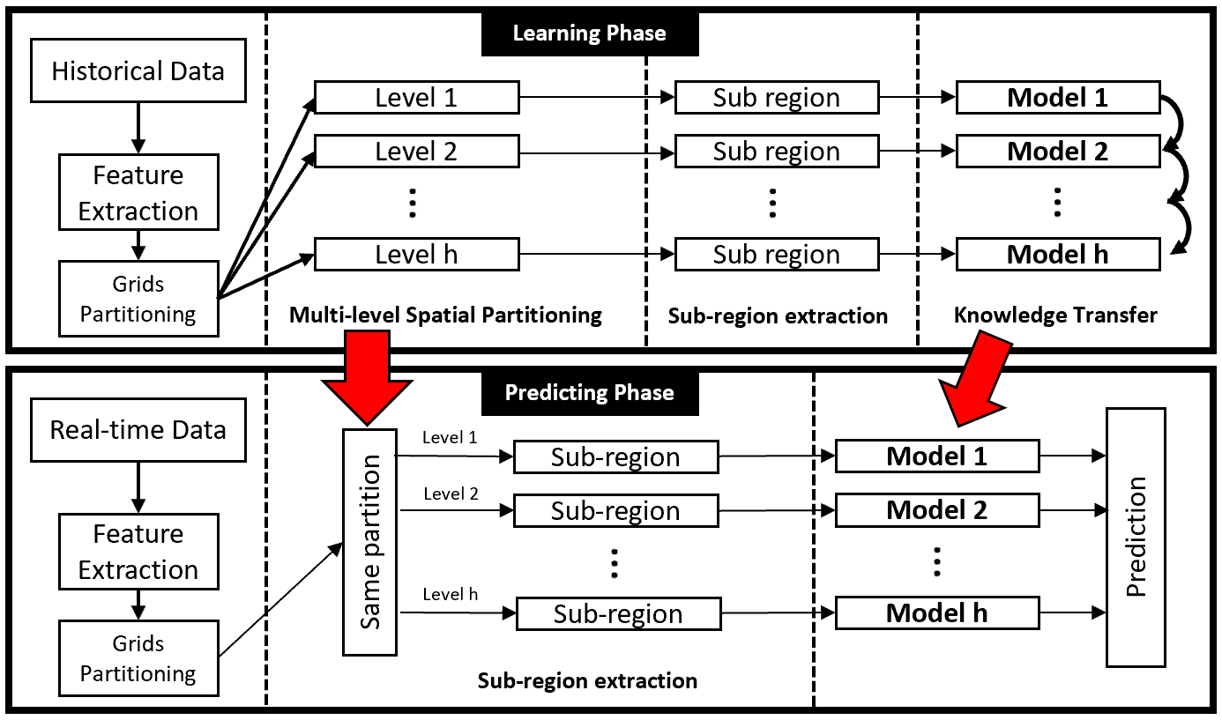}
	\vspace{-0.1in}
	\caption{The HintNet Solution Overview.}
	\vspace{-0.1in}
	\label{overview}
\end{figure}

\vspace{-0.1in}
\section{Proposed Solution}
In this section, we present our solution HintNet. Figure \ref{overview} shows the overview of the proposed framework. 
In the learning phase, we first perform the Multi-level spatial partitioning to obtain levels of regions with different risks. Meanwhile, the sub-regions are extracted and used to train models on each risk level. Lastly, the knowledge learned from previous models is transferred to the next level by initializing model parameters. In predicting phase, the real-time data is extracted and mapped into grids, then we use partitioned results from the training phase to classify grids into the same levels. Finally, features are fed into corresponding well-trained models to make final predictions.
\vspace{-0.1in}
\subsection{Multi-level Risk-Based Spatial Partitioning}
To address the limitations of related work in capturing irregular spatial heterogeneity patterns, we propose a spatial partitioning method, namely, Multi-level Risk-Based Spatial Partitioning (M-RSP) to partition the grids into irregular-shaped regions based on the accident risk in a hierarchical manner. 
Specifically, M-RSP applies binary-partitioning iteratively. In each step, the study area is split into a risky region and less-risky region. In the next iteration, M-RSP repeats this binary-partitioning process on the previous risky region. By doing this iteratively, we obtain multiple levels of partitioned less-risk regions. Each level of less-risk regions represents a hierarchy of the risk distribution. To distinguish risk levels, we use a threshold $\eta$ to compare with the total number of accidents in each partitioned region. If the accident count is greater than $\eta$, this region is risky. Otherwise, the region is less-risky.

The partitioning procedure for every single level, which we refer to as RSP (Risk-Based Spatial Partitioning), is inspired by DBSCAN \cite{10.5555/3001460.3001507}, and designed for binary-partitioning on grids with similar risk. Given a cell location $l$, its \textbf{neighbors} are defined as cells within a \textbf{range $\epsilon$} both horizontally and vertically in Manhattan distance. \textbf{Min\_points $\gamma$} is a threshold used to identify grid cell as \textbf{high-risk}, if its accidents count is greater than $\gamma$. \textbf{Min\_Risk $\lambda$} is used for filtering out noise cells within fewer accidents than this limit. A \textbf{critical cell} is defined as a cell with the number of high-risk neighbors including itself greater than a threshold \textbf{min\_neighbors $\beta$}. A \textbf{border cell} is a non-critical cell with at least one critical cell in the neighbors. An \textbf{outlier} is not a critical cell and has no critical cells among its neighbors. In RSP, grid cells are classified as Critical cells, Border cells, or Outliers. 
Specifically, for each level, RSP starts with checking a random cell. If this cell is identified as a Critical cell, and then its neighbors will also be checked until no Critical cell is identified. The border cell and critical cell are assigned with a partition label. The algorithm repeats this process to the next random unclassified grid cell until all grids are classified. Importantly, We have this unique design of identifying Critical cells by counting their high-risk neighbors. The reason behind this design is that if we simply count the total accidents within a region, some low-risk grid cells with an extremely high-risk neighbor cell will be classified as high-risk cells as a result of over-influence from that high-risk neighbor. Lastly, grid cells with no road are filtered out by using the mask map $H$. 

\begin{figure*}
    \centering
	\includegraphics[width=1\textwidth]{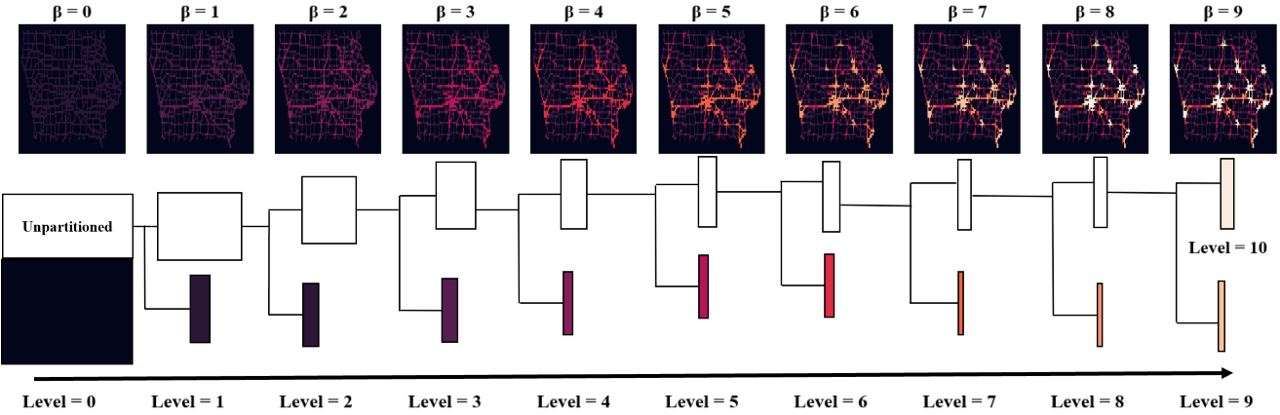}
	\vspace{-0.1in}
	\caption{\textbf{Multi-level partitioning}. White box - unpartitioned regions that will be checked in the next level. Colored box - partitioned regions, size indicating the number of partitioned grid cells in corresponding level.}
	\vspace{-0.1in}
	\label{Multi-level}
\end{figure*}

The overall M-RSP algorithm calls for the single-level RSP procedure iterative for partitioning on each level for $\beta$ iterations until all the levels are generated. Algorithm 1 shows the details of the entire M-RSP, where Line 1 determines the maximum number of iterations, which equals to maximum neighbors controlled by $\epsilon$. Line 5 performs the partitioning for a single level using RSP. Line 6 checks for the last iteration to partition all left regions by setting $\eta$ as infinity. The remaining algorithm implements binary-partitioning by checking with $\eta$. In the end, the result is incremented by one because the noise level was initialized as negative ones. 
\begin{algorithm2e}[t]
	\SetKwInput{Input}{Input}
	\SetKwInput{Output}{Output}
	\LinesNumbered
	\DontPrintSemicolon
	\BlankLine
	\caption{Multi-level Risk-based Spatial Partitioning (M-RSP)}
	\label{pseudo_learn}
	\Input{accidents matrix $A$, mini\_points $\gamma$, min\_risk $\lambda$, threshold $\eta$, epsilon $\epsilon$}
	\Output{matrix with assigned level label}
	\BlankLine
    $iter = (2*\epsilon + 1)^2$ \; 
    
    Initialize $zero$ map matrix $z$ \;
    Initialize $result$ as $-1$ map matrix \;
    \For{each $\beta$ from 0 to iter}{
        $Partitions =$ RSP$(A, \epsilon, \gamma,  \lambda, \beta)$\;
        \If{ $\beta$ == iter }{
            $\eta = +\infty$
        }
        \For{each p in $Partitions$}{
            \For{each g in p}{
                \If{$p[g] == 0$ and $ z[g] == 0$}{
                    $z[g] = 1$\;
                    $result[g] = \beta - 1$\;
                }
                }
            $ctr$ = CountAccidents($p$, $A$)\;
            \If{$ctr$ $<=$ $\eta$}{
                \For{each g in p}{
                    \If{$z[g]$ == 0 }{
                        $z[g]$ = 1\;
                        $result[g] = \beta$\;
                    }
                }
            }
            
        }
    }
    $result$ += $1$ \;
	\textbf{return} $result$
\end{algorithm2e}
Figure \ref{Multi-level} illustrates this process of multi-level partitioning on the entire Iowa dataset. 
In the binary tree, levels represent partitioned maps with $\beta$ from 0 to 9 when $\epsilon$ is set as $1$. The parent node represents the unassigned cells from the previous level. The left child represents the assigned grids in the current level, and the right child node represents unassigned grids. The right-most map on the top is the final risk-based partitioning map. Lastly, depending on the granularity of the partition result we need, we can aggregate every $k$ levels together into a single partition, where $k$ is a tunable hyperparameter in our framework. The finest granularity is when $k=1$, which is the same as using the original partitioned result.
\vspace{-0.1in}
\subsection{HintNet deep learning solution}
\begin{figure}
    \centering
	\includegraphics[width=0.45\textwidth]{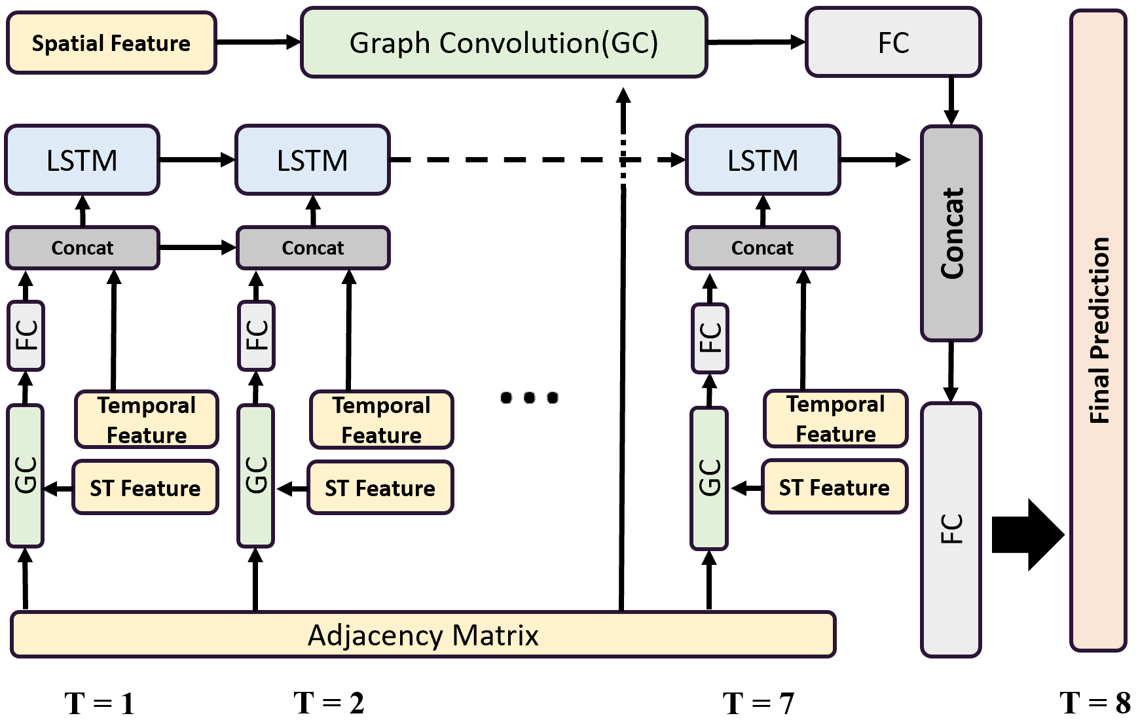}
	\vspace{-0.1in}
	\caption{HintNet Deep Learning framework.}
	\vspace{-0.23in}
	\label{DL_model}
\end{figure}

Given the partitioned regions on each level of the hierarchy, separate models are trained on each level to reduce the spatial heterogeneity issue. However, simply applying separate models ignores the potential connectivity between regions on each level of the hierarchy. Thus, we develop a knowledge transfer mechanism to allow models to share learned knowledge across different levels. 
Figure \ref{DL_model} shows the basic structure of HintNet. The inputs include spatial features, temporal features, spatial-temporal features, and an adjacency matrix from the same region. The output of HintNet is the predicted number of accidents. Firstly, The Graph Convolution(GC) is carefully designed to capture the local spatial auto-correlations
along with the road system. Meanwhile, the training process should remain reasonably efficient. To achieve that goal, sub-regions are extracted from each level. In every time interval $t$, we treat each cell $l$ and its surrounding neighbor cells as one $w \times w$ image, where the size $w$, a hyperparameter, controls the filter size of Graph Convolution. 
The sub-region grids are converted into a graph, where cells are treated as vertices and correlations between cells are treated as edges. Inspired by the dynamic CNN \cite{8970742} proposed by Zhang et al., We use the Pearson correlation coefficient in Equation \ref{Pearson} to quantify the accident correlations between grid cells to capture the spatial dependencies. 
\begin{equation}\label{Pearson}
a_{XY} = \frac{\sum_{i=1}^n (X_{i}-\overline{X})(Y_{i}-\overline{Y})}{\sqrt{\sum_{i=1}^n (X_{i}-\overline{X})^2 \sum_{i=1}^n (Y_{i}-\overline{Y})^2}}
\end{equation}
where the adjacency matrix $A$ is a symmetric $w^{2} \times w^{2}$ matrix such that its element $a_{XY}$ is the accident correlation between location $X$ and $Y$.  Thus, for each cell $l$ and time interval $t$,  a feature image tensor $X^t_{l}\in \mathbb{R}^{w \times w \times d}$ is retrieved, where $d$ is the dimension of the features. The inputs of the Graph Convolution layer are $X^t_{l}$ for each cell $l$ and corresponding adjacency matrix $A$, and the output is output feature matrix $H_{l}$.
\begin{equation}
H^{i}_l = f(H^{i-1}_l, A) = \sigma(A H_l^{i-1}W_{i}), 
\end{equation}
Where $H^{i}_{l}$ is the output feature matrix of the region $l$ in the i-th layer. In this way, Graph Convolution filters focus more on regions with higher correlations along with the road network and ignore the regions with irrelevant traffic accident patterns.

Secondly, we use Long Short-Term Memory(LSTM) \cite{HochreiterSepp1997LSM} as building blocks. To capture the spatial dependencies, Graph Convolution layers are first applied on spatial-temporal features to obtain filtered feature matrix $H_l$ in each time interval $t$. Afterward, we concatenate the last output feature matrix $H_l$ with temporal features and feed it into a fully connected layer, and then the output is used as input of each LSTM state. Concurrently, Spatial Features are also fed into a Graph Convolution layer, and another fully connected layer is applied on its output feature matrix to get low dimension representation. Lastly, the representation of spatial features is concatenated with the output of the last LSTM state, and then we fed the concatenation into the last fully connected layer to make final predictions.

\textbf{Cross-level Knowledge Transfer}
We argue that there exist some common traffic accident patterns in all levels of hierarchy, especially for particular adjacent levels. For example, the risk factors recognized in downtown areas such as holiday events can be transferred to nearby regions, because they tend to share similar traffic patterns. To transfer learned knowledge across levels, we transfer model parameters learned from the previous level to the model in the next level. Compared with initializing model parameters randomly, we argue that the transferred parameters learned from other levels carry the experience of forecasting accidents and contribute to the training process of other levels. The right part of Figure \ref{overview} shows this process. In our case, the knowledge is transferred from the level of urban areas to the level of rural areas (i.e., from leaf to root).
During the training phase, we apply the gradient descent to update parameters $\theta$, with a learning rate $\alpha$. The training process uses the mean square error (MSE) as a loss function.
\vspace{-0.10in}
\begin{equation}
Loss =  \frac{1}{T} \sum^T_{t=1} (Y_t - \hat{Y}_t)^2,
\vspace{-0.15in}
\end{equation}
Where $Y_t$ is the ground truth and $\hat{Y}_t$ is the predicted values of all grid cells at time interval $t$.\\
Algorithm \ref{Transfer} shows the training process of the cross-level knowledge transfer. The output contains trained models on each level, and they are used in Predicting Phase. In the Predicting Phase, real-time features in each risk level are fed into corresponding trained models and make predictions on each level. The final prediction of HintNet is the integration of predictions on each level. 

\begin{algorithm2e}
	\SetKwInput{Input}{Input}
	\SetKwInput{Output}{Output}
	\LinesNumbered
	\DontPrintSemicolon
	\BlankLine
	\caption{Cross-level Training}
	\label{Transfer}
	\Input{Multi-level partitioning $\tau$, predictor $f()$,
	True accidents $Y$, learning rate $\alpha$, Epoch $e$} 
	\Output{trained models on each level}
	\BlankLine
    initialize parameter $\theta$ in $pool$ for each level\;
    
    \For{each level $v$ in $\tau$}{
    
    \If{not 1st v}{
            $\theta$ = $pool[$previous $v]$
        }
        
    \For{each iteration in $e$}{

        $F_T, F_S. F_{ST}, A$ = SubregionExtract($v$) \;
        $\hat{Y}$ = $f_\theta(F_T, F_S. F_{ST}, A)$ \;
        $Loss$ = $\frac{1}{T} \sum^T_{t=1} (Y_t - \hat{Y}_t)^2$\;
        Calculate gradient $\nabla g(\theta)$ by $Loss$\;
        Update $pool[v]$ = $pool[v]$ + $\alpha \nabla g(\theta)$ \;
        }
        }
	\textbf{return} $pool$
\end{algorithm2e}

\vspace{-0.1in}
\section{ Evaluation}
In this section, we demonstrate the effectiveness of our proposed method comparing with baselines.
\vspace{-0.1in}
\subsection{Experiment Settings} 
In this part, we will explain the basic setting of our experiments.\\
\textbf{Data Preprocessing: } 
We use the data from the first two years (2016-2017) as a training set, and the validation set is randomly selected from $20\%$ of the training set. The data from the year 2018 is used as a testing set. Besides, the whole state of Iowa area is partitioned into 5km by 5km grids.

\textbf{Evaluation Goals: }(1) Does the proposed framework outperforms baseline methods with different levels of heterogeneity? (2) Which features have the most impact on prediction accuracy in different levels. (3) Which granularity $k$ of partitioning results in the best performance? (4) Does the knowledge transfer mechanism improve the model performance? 

\textbf{Metrics:}
We evaluate the performance of the models by measuring the mean square error (MSE) Besides, we use the number of model parameters to demonstrate the resources usage between the proposed model and baselines.

\textbf{Baselines: }
We compare our proposed framework with the following baselines: (1) Least Square Linear Regression
(2) FC-LSTM, two layers of LSTM  (2) Decision Tree Regression. The max depth is set to 30. (3) ConvLSTM \cite{10.5555/2969239.2969329}. We use a two-layer structure and the hidden dimension is equal to the number of features. (4) Hetero-ConvLSTM. Each ConvLSTM in hetero-ConvLSTM uses the same parameter setting as the ordinary ConvLSTM baseline. We use multiple moving windows with fixed size $32 \times 32$. The number of windows depends on the size of the study region. (5) GSNet. we modify the weighted loss function part to fit into our case. (6) Historical Average. Historical average daily accident counts. The mask map is also applied on all baseline predictions to filter out cells without roads. 
\vspace{-0.3in}

\subsection{Performance Comparison} 
\begin{figure}
    \centering
	\includegraphics[width=0.4\textwidth]{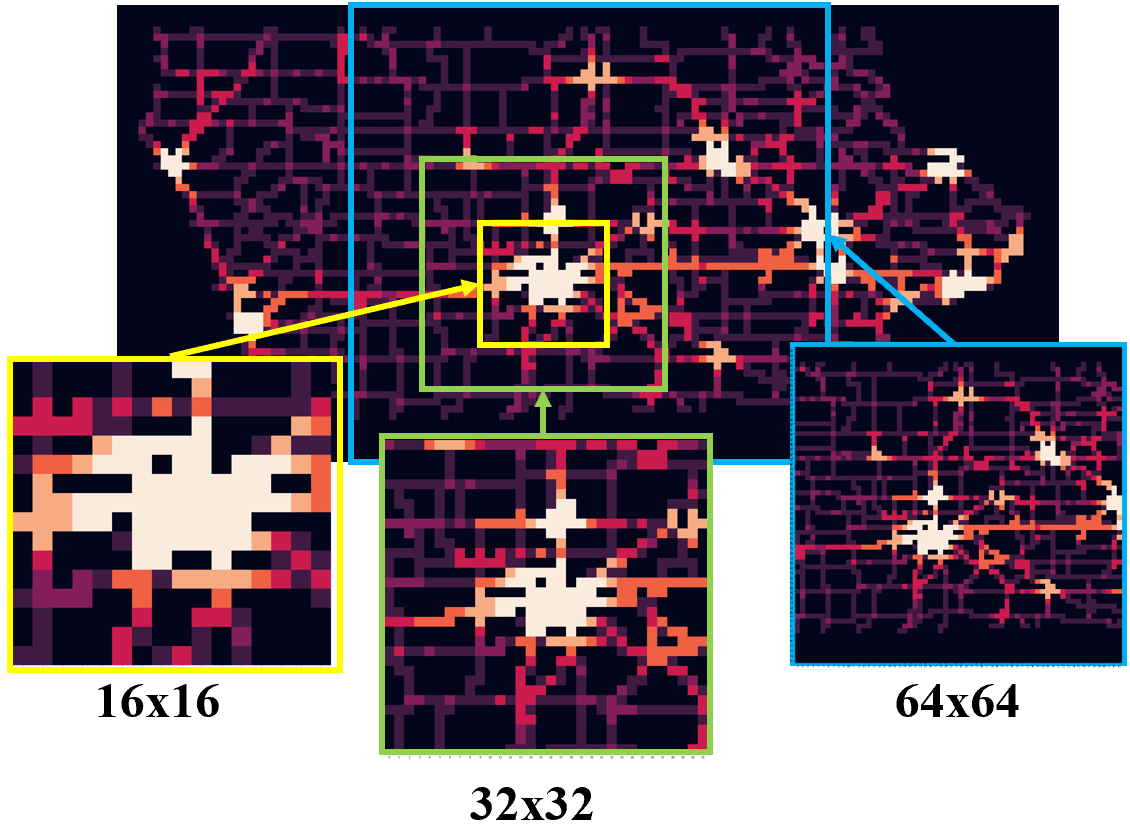}
	\vspace{-0.1in}
	\caption{Illustration of four test regions}
	\vspace{-0.3in}
	\label{subregion}
\end{figure}

We compare the performance between the proposed method and baselines given with different levels of spatial heterogeneity. 
We test the models in 4 types of regions with grid size $16 \times 16$, $32 \times 32$, $64\times64$, and $128\times64$ separately. Figure \ref{subregion} shows the corresponding grid map representing homogeneous, less-homogeneous, heterogeneous, severely heterogeneous regions respectively. To test all methods on them, the filter size of ConvLSTM and the sub-region size in our method is set as $5 \times 5$ in the homogeneous region ($16 \times 16$) and less-homogeneous region ($32 \times 32$) grid map. For heterogeneous region ($64 \times 64$) and severe-heterogeneous region ($128 \times 128$), the both sizes are set as $7 \times 7$. Besides, GSNest is infeasible to be applied on the largest region ($128 \times 64$) because of its oversized graph convolution part. Therefore, we applied two independent GSNet models on the left-half and right-half parts of the study area, and calculate the total prediction error. For Hetero-ConvLSTM, the number of moving windows used for each region are 1, 1, 9, 21 respectively, as a result of its fixed window size.

Table \ref{baseline comparison} compares models in all heterogeneity levels, where HintNet outperforms all baseline methods. State-of-the-art methods like GSNet and ConvLSTM achieve decent prediction in smaller regions with less spatial heterogeneity but perform poorer in large, heterogeneous regions. 
%
In this case, HintNet outperforms GSNet $43\%$. On the opposite, our method and hetero-ConvLSTM address the heterogeneity problem and perform stably in all regions. Nevertheless, Hetero-ConvLSTM suffers from an excessive number of sub-models, square-shaped partitioning with a fixed size, and dis-connectivity between sub-models. In a less-homogeneous region, our method outperforms hetero-ConvLSTM by $26\%$ in prediction error. 
\textbf{Model Complexity:} The experiments show that HinNet achieves better performance with a relatively smaller model size. The results of model complexity are included in supplementary materials. The supplementary material and code is available at: \url{https://github.com/BANG23333/HintNet}.

\begin{table*}[]
	\centering
	\caption{Model Comparison}
	\begin{tabular}{|c|c|c|c|c|c|c|c|c|}
		\hline
    		& \textbf{LR} & \textbf{DTR} & \textbf{LSTM} & \textbf{ConvLSTM} & \textbf{GSNet}  & \textbf{Hetero-ConvLSTM} & \textbf{HA} & \textbf{HintNet} \\
		\hline
		\hline
		$16 \times 16$ & $0.221$ & $0.388$ & $0.264$ & $0.197$ & $0.179$ & $0.197$ & $0.220$ & $\textbf{0.174}$ \\
		\hline
		$32 \times 32$ & $0.075$ & $0.133$ & $0.085$& $0.081$ & $0.080$ & $0.081$ &$0.074$ & $\textbf{0.060}$ \\
		\hline
		$64 \times 64$ & $0.032$ & $0.085$ & $0.033$ & $0.044$ & $0.036$ & $0.028$ &$0.032$ & $\textbf{0.025}$\\
		\hline
        $128 \times 64$ &$0.026$ & $0.033$ & $0.027$ & $0.039$ & $0.037$ & $0.024$ & $0.027$ & $\textbf{0.021}$\\
        \hline
	\end{tabular}
	\label{baseline comparison}
\end{table*}

%

\vspace{-0.1in}
\subsection{Ablation study}
In this section, we first evaluate HintNet's performance on each level of the hierarchy, then compare the impact of features and parameters.

\begin{figure}
    \centering
	\includegraphics[width=0.4\textwidth]{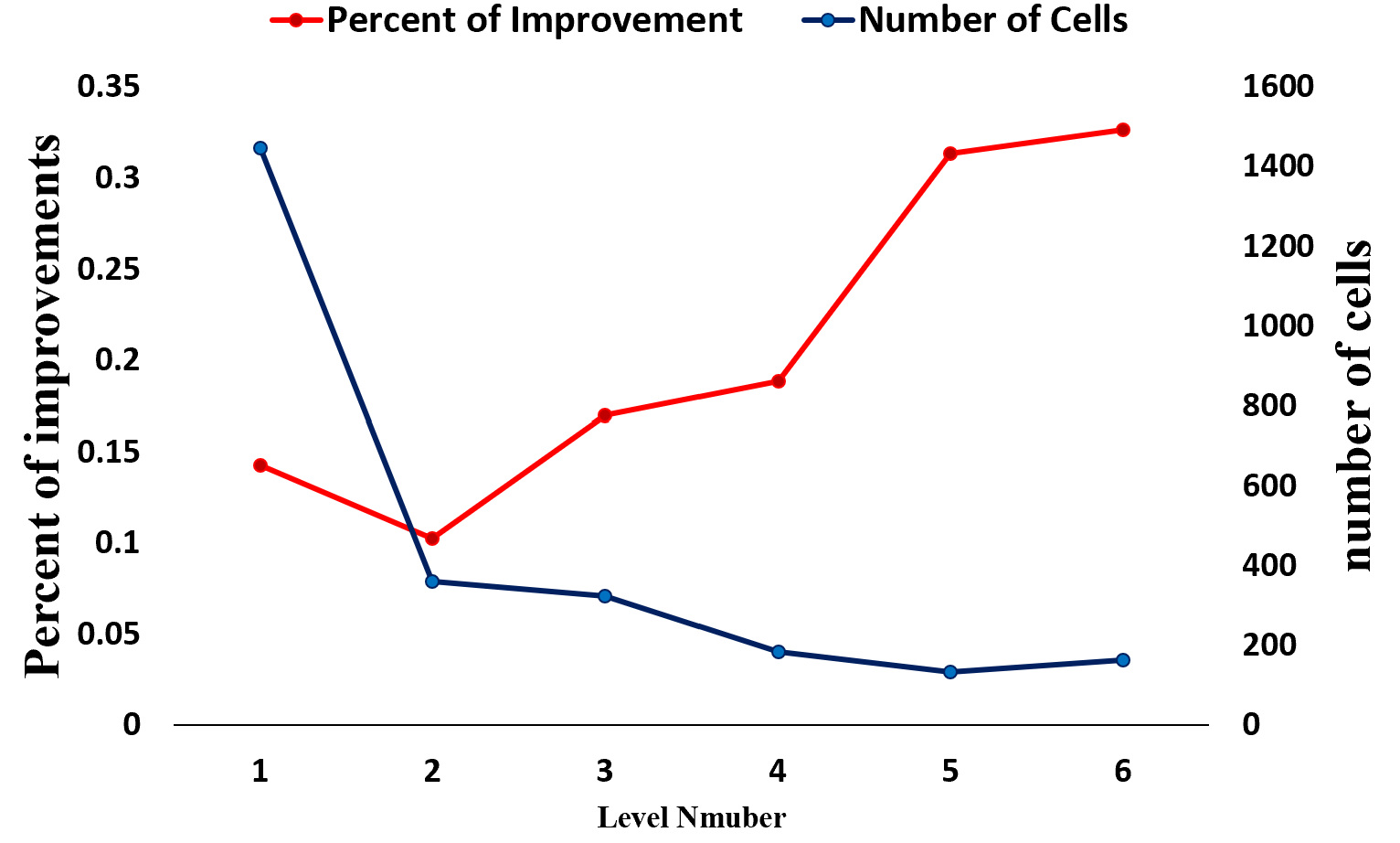}
	\vspace{-0.1in}
	\caption{Improvement across levels}
	\vspace{-0.3in}
	\label{trend}
\end{figure}

\textbf{Improvements on each level:}
We study how our proposed model performs in different levels of the partition.
The Historical Average(HA) is generally a good reference value to measure the overall improvements of HintNet in each level. We calculate the percent of improvements between our proposed model and the historical average in each level. Figure \ref{trend} shows the trend of improvements from rural regions to urban regions. The red line represents the percentage of improvement and the blue line represents the number of grid cells involved in each level. As we can see, the improvements grow steadily as the level number grows. This indicates that HintNet makes relatively better predictions in urban areas and suburban areas, but only slightly exceeds HA in rural areas partially due to data sparsity.

\textbf{Impact of feature group: }
To determine the effectiveness of different feature groups on the results, we examine the results by adding feature groups one by one. 
As Table \ref{features} illustrated, the model with only spatial features (S) has slightly better performance than the historical average. 
With extra temporal features (T) such as calendar data, our model makes a great improvement on level 6 which represents the downtown areas. It reveals that calendar features enable the model to capture the temporal patterns in areas with frequent human activities such as traffic jams in holidays. Interestingly, the spatial-temporal (ST) features like weather information brings down the errors on level 5, level 6, and level 4 significantly. This indicates that the dynamic weather changes play a significant role in predicting accidents in state highways and residential areas which covers the large road systems with high-speed traffic volumes. 
Lastly, the prediction on extreme rural areas including level 2 and level 1 is still challenging due to the randomness of accidents.
\begin{table}[]
	\centering
	\caption{Error Comparison with different k}
		\vspace{-0.1in}
	\begin{tabular}{|c|c|c|c|}
		\hline
    		& \textbf{k = 3} & \textbf{ k = 2} & \textbf{k = 1} \\
		\hline
		\textbf{MSE} & $0.063$ & $0.021$ & $0.135$ \\
		\hline
	\end{tabular}
	\label{levelbylevel}
\end{table}

\begin{table*}[]
	\centering
	\caption{Impact of feature groups}
	\vspace{-0.1in}
	\begin{tabular}{|c|c|c|c|c|c|c|c|}
		\hline
    		& \textbf{level 6} & \textbf{level 5} & \textbf{level 4} &\textbf{level 3} &\textbf{level 2} &\textbf{level 1} &\textbf{All levels}\\
		\hline
		\hline
		\textbf{S} &$0.676$ & $0.164$ &$0.098$ & $0.057$ & $0.036$ & $0.007$& $0.024$\\
		\hline
		\textbf{S+T} &$0.648$ & $0.150$ &$0.090$ &$0.053$ & $0.036$ & $0.007$& $0.022$\\
		\hline
		\textbf{S+T+ST} &$0.625$ & $0.118$ & $0.066$ &$0.049$ & $0.034$ & $0.007$& $0.021$\\
		\hline
	\end{tabular}
	\label{features}
	\vspace{-0.18in}
\end{table*}

\begin{figure}
    \centering
	\includegraphics[width=0.41\textwidth]{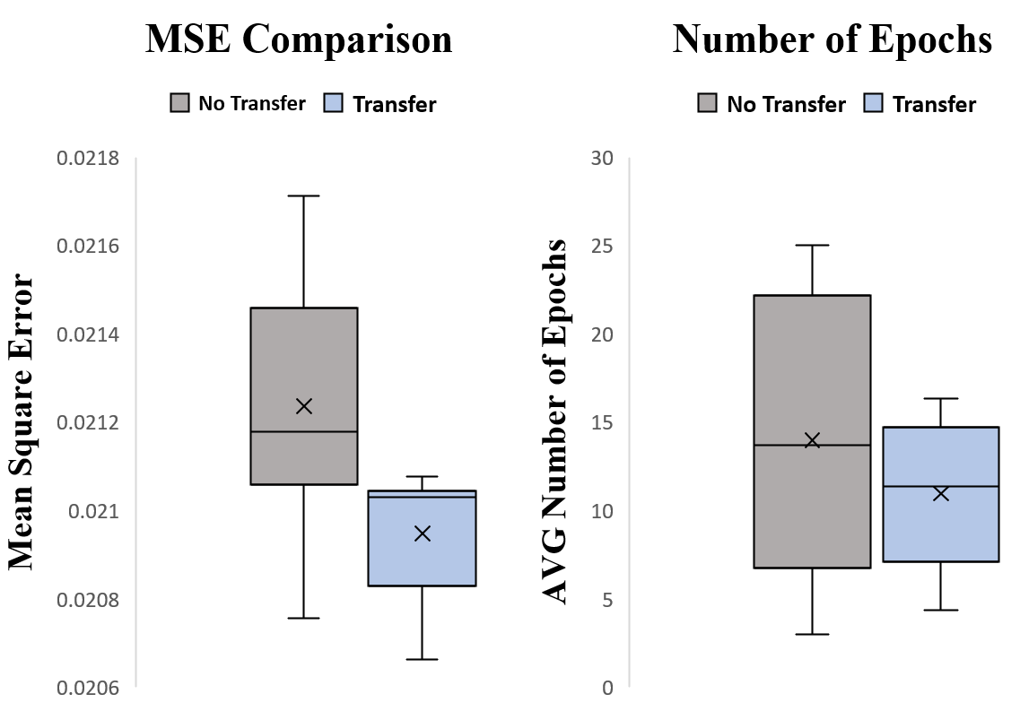}
	\vspace{-0.1in}
	\caption{Knowledge Transfer}
	\vspace{-0.2in}
	\label{transfer}
\end{figure}

\textbf{Impact of risk-tree level aggregation granularity $k$: }
To investigate which degree of partitioning granularity leads to best model accuracy. We test the model when $k$ is set to be 3, 2, and 1. As shown in Table \ref{levelbylevel}, when $k=2$ we obtain the best result. When the $k=3$, there are only 3 partitioned levels. The cells with different accident patterns are mixed in a single group, so the model cannot address the spatial heterogeneity appropriately. Similarly, the model has the worst results when $k=1$. In this case, the over-fined granularity results in a limited number of samples in each level, and it becomes extremely difficult for our model to learn from levels with more variability.

\vspace{-0.1in}
\subsection{Impact of cross-level knowledge transfer}
We assume that the inherent knowledge learned from levels helps the training process on other levels. We examine the benefits of using the knowledge transferring mechanism based on its accuracy and efficiency. To check the improvements in prediction accuracy, we train our models with and without knowledge transfer mechanism under the same parameter setting for $10$ times, and then we draw a box-plot from their mean square errors of testing sets. Figure \ref{transfer} shows that models with knowledge transfer have lower average and lower variance. On the other hand, to test the improvements on model efficiency, we first use the training set and validating set to get the best validating errors on each level and use them as target lines. Next, we train the models with and without knowledge-transferring for $10$ times to reach the same target validating errors on each level. We record the average number of epochs the models take to reach the target error. Lastly, we draw a box plot based on their level-average epoch cost. We can find that models with knowledge transfer mechanism have less average epoch cost and less variance. The experiments results show that knowledge transfer can further improve and stabilize prediction accuracy and training efficiency.

\vspace{-0.1in}
\section{Conclusion}
 In this paper, we performed a comprehensive study on the traffic accident forecasting problem. Traffic accident prediction is important to transportation management and public safety, but it is very challenging due to spatial heterogeneity and rareness. We proposed a HintNet model to partition areas into multi-level sub-regions based on their accident risk and learn models for each level. Meanwhile, A knowledge transfer mechanism is applied across different levels. The experiments show that the HintNet is a promising solution to accident prediction problems, and HintNet outperforms the state-of-art method up to 12.5$\%$ on prediction error.
 
 \vspace{-0.1in}
 \section{Acknowledgements}
 Bang An and Xun Zhou are partially supported by an ISSSF grant from the University of Iowa and the SAFER-SIM UTC under US-DOT award 69A3551747131. Yanhua Li was supported in part by NSF grants IIS-1942680 (CAREER), CNS1952085, CMMI-1831140, and DGE-2021871.

\vspace{-0.1in}
\bibliographystyle{plain}
\bibliography{references}

%
%




\end{document}